\documentclass[numbered]{trbunofficial}
\usepackage{graphicx}

\usepackage[hidelinks]{hyperref}
\usepackage{tabularray}
\usepackage{tabularx}

\usepackage{rotating}
\usepackage{amsmath}
\usepackage{subfig}

\AuthorHeaders{Kamal, Farooq}
\title{Copula-ResLogit: A Deep-Copula Framework for Unobserved Confounding Effects}

\author{%
  \textbf{Kimia Kamal}\\
  Laboratory of Innovations in Transportation (LiTrans)\\
  Toronto Metropolitan University \\
  Toronto, Canada \\
  Email: kimia.kamal@torontomu.ca\\
  \hfill\break
  \textbf{Bilal Farooq}\\
  Laboratory of Innovations in Transportation (LiTrans) \\
  Toronto Metropolitan University \\
  Toronto, Canada \\
  Email: bilal.farooq@torontomu.ca\\
  \hfill\break%
}

\usepackage{adjustbox}

\usepackage{caption}

\usepackage{tabularray}
\UseTblrLibrary{booktabs}

\usepackage{tabularx}
\usepackage{multirow}

\begin{document}
\maketitle

\section{Abstract}
A key challenge in travel demand analysis is the presence of unobserved factors that may generate non‑causal dependencies, obscuring the true causal effects. To address the issue, the study introduces a novel deep learning based fully-interpretable joint modelling framework, Copula‑ResLogit, which integrates the flexibility of Residual Neural Network (ResNet) architectures with the dependence capturing capabilities of copula models. This hybrid structure enables us to first detect unobserved confounding through traditional copula‑based joint modelling and then mitigate these hidden associations by incorporating deep learning components. The study applies this framework to two case studies, including the relationship between stress levels and wait-time of pedestrians when crossing mid-block in VR and the dependencies between travel mode choice and travel distance in London travel behaviour data. Results show that Copula‑ResLogit substantially reduces or eliminates the dependencies, demonstrating the ability of residual layers to account for hidden confounding effects.

\hfill\break%
\noindent\textit{Keywords}: Travel behaviour, joint choices modelling, copulas, causal learning, residual neural network
\newpage

\section{Introduction}
The process of decision-making regarding travel choices is influenced by multiple factors---humans view multiple variables as an integrated system and make their decisions based on inter-relationships between variables \cite{bhat1997work,kuppam2001structural}. Indeed, there may exist shared attributes that simultaneously influence multiple factors, introducing dependency among them. These shared variables, whether observable or unobserved, can lead to non-causal associations among the dependent variables in behavioural studies. In the context of travel behaviour analysis, accounting for potential non-causal relationships between variables within the decision-making process leads to more robust and unbiased insights for assessing transportation systems and informing effective policy-making. In fact, in what-if analysis scenarios, it is essential to adopt modelling approaches that more accurately represent the real-world decision-making process and distinguish the direct effects between dependent variables from non-causal associations.

In statistics, joint modelling is an approach that involves analyzing the dependency or correlation between multiple variables. This approach is able to capture dependencies between factors, leading to more accurate travel choice analysis. The copula approach is a technique used within the realm of joint modelling to capture the dependency structure between random or unobserved variables without imposing distribution assumptions, characterizing them as a dependent multivariate distribution \cite{bickel2009springer, trivedi2007copula}. The copula method has emerged as a powerful tool in transportation studies for measuring dependencies among multiple factors influencing a travel decision. This method uncovers both the direction and magnitude of the correlations among dependent variables. Therefore, the copula-based approach enables us to identify the impact of unobserved confounding and non-causal associations between variables. The fundamental idea underlying this approach is that dependent variables may share unobserved factors, which in turn induce statistical dependence between them. It is important to note that, according to principles of causal inference, this form of dependence differs from a direct causal association between the dependent variables \cite{brathwaite2018causal, pearl2009causal}. Generally, the copula framework has been classified under the theory-driven methods, involving the specification of a priori functional forms for each dependent variable or component of the decision-making process \cite{spissu2009copula, bhat2009copula}.

In recent years, the impressive capabilities of Machine Learning (ML) and Deep Neural Networks (DNNs) in providing more flexible model structures with higher prediction accuracy have made them promising alternatives to traditional travel behaviour analysis. Despite the ability of data-driven approaches to capture nonlinear relationships and enhance predictive accuracy, their architecture may also introduce certain limitations. In the context of discrete choice modelling, there has been a surge of interest regarding the combination of Random Utility Maximization models (RUMs) with deep learning algorithms \cite{wang2020deep, wong2021reslogit, kamal2024ordinal, martin2021revisiting, sifringer2020enhancing}. The primary objective of these hybrid deep discrete choice models is to address the limitations inherent in both traditional discrete choice models and machine learning approaches. For instance, studies such as \citep{wong2021reslogit} and \citet{kamal2024ordinal} focused on the lack of interpretability in machine learning models and the restrictive assumptions of model specifications in traditional categorical and ordinal discrete choice frameworks. In general, hybrid data-driven approaches aim to improve prediction accuracy while making the two modelling frameworks complementary. This study also employs a hybrid data-driven approach, albeit with a different focus. Drawing on insights from previous research, we investigate the potential of deep learning to eliminate unobserved confounding effects through a hybrid modelling framework.

The main objective of this research is to leverage the complementary strengths of data-driven and theory-driven models for: 1) investigating potential non-causal dependencies between two factors arising from shared unobserved variables, and 2) examining the power of deep neural networks to mitigate these dependencies in support of causality-based decision-making. In other words, not only does this research aim to explore and highlight the correlation between dependent variables in the choice selection process stemming from unobserved variables, but it also tries to examine the power of deep learning to control for unobserved confounding associations. To the best of our knowledge, this is the first effort in the literature to investigate this objective through a hybrid modelling approach. This study proposes a hybrid data-driven copula-based joint behavioural model aimed at evaluating the capacity of deep learning algorithms to disentangle causal associations from non-causal associations arising due to unobserved confounding factors. The proposed model is built by integrating the copula-based joint modelling approach with ResNet-based model structures. ResNet-based model structures, including ResLogit and Ordinal-ResLogit, are deep interpretable learning discrete choice frameworks used for both ordinal and categorical datasets \cite{wong2021reslogit, kamal2024ordinal}. This hybrid proposed model is referred to as Copula-ResLogit in this study. In addition to the main objective of this study, the proposed model also addresses a key limitation of copula-based joint approach in the functional form of models, which must be specified in advance of model estimation. Furthermore, consistent with the hybrid strategies adopted in \citep{wong2021reslogit} and \citep{kamal2024ordinal} studies, our hybrid approach preserves the interpretability and the consistency of the model with expert knowledge. It is worth noting that while the primary focus of this study lies within the realm of discrete choice modelling, the proposed model possesses the flexibility to extend its applicability to joint discrete/continuous modelling as well. 

In this study, we utilize both Stated Preference (SP) and real-world Revealed Preference (RP) datasets to evaluate the performance of the Copula-ResLogit model in causality analysis studies \cite{hillel2018recreating, mudassar2021analysis}. First, two key components of the pedestrian crossing decision process—stress level and waiting time—are jointly analyzed using a Virtual Reality (VR) dataset. This dataset employs a novel data collection approach, in which participants are immersed in artificial, computer-generated environments \cite{farooq2018virtual}. Secondly, the RP London Travel Demand dataset is utilized to investigate potential dependencies between travel distance and travel mode choice that may stem from unobserved factors, influencing travel behaviour decisions.

The key contributions of this study to causality-based travel behaviour modelling literature are as follows:

\begin{itemize}
    \item Measures and captures confounding associations between dependent variables rooted in unobserved variables.
    \item analyzes the power of deep learning to control for unobserved confounding in causal inference.
    \item enhances the predictive accuracy of the model
    \item Retains the capability of behavioural indicators analysis.
\end{itemize}

The remainder of this paper is organized as follows. First, a brief overview of the copula approach and its application in transportation studies is presented. Second, the methodology section explains the structure of the Copula-ResLogit and theory-driven copula models for different problems. This is followed by a description of the datasets. Section 4 provides an analysis of the estimated models. Finally, the conclusion section offers suggestions and future directions for the study.

\section{Background}
\subsection{Copulas in transportation} \label{Copulas in transportation}

Coupla functions associate the distribution function of several random variables through their marginal distributions and define them as a dependent multivariate distribution \cite{trivedi2007copula, nelsen2006introduction}. The copula approach is based on Sklar’s theorem saying that for multiple random variables $X_{1}, X_{2}, \dots, X_{n}$  with marginal distribution functions $F_{1}, F_{2}, ..., F_{n}$ and joint distribution function $H$, there exists a n-dimensional copula function C \cite{trivedi2007copula, nelsen2006introduction}. This can be written as follows:

\begin{equation}
\label{Sklarstheorem}
H(x_{1},x_{2}, ...x_{n})= C (F_{1}(x_{1}),F_{2}(x_{2}), \dots, F_{n}(x_{n}))
\end{equation}

To clarify this equation, for two random variables $X_{1}, X_{2}$, each pair of real numbers $x_{1}$ and $x_{2}$ can be mapped to a pair $((F_{1}(x_{1}),F_{2}(x_{2}))$ in the unit square space $[0,1]^2$, and this pair corresponds to a joint distribution $H(x_{1},x_{2})$ within the range [0,1]. Therefore, according to Equation \ref{Sklarstheorem}, the copula function is a function that links the joint distribution of multiple random variables to their marginal distributions. 


    

Based on the definition of joint distribution, the n-dimensional copula for $X_{1}, X_{2}, \dots, X_{n}$ random variables can be written as:

\begin{equation}
\label{Joint dist}
 C_\theta (F_{1}(x_{1}),F_{2}(x_{2}), \dots, F_{n}(x_{n}))= Pr(X_{i} \leq x_i ; i = 1,2, ..., n)
\end{equation}

\noindent where $\theta$ is the copula parameter vector measuring the dependency between the random variables. In copula-based joint modelling, the $m(m-1)/2$ dependency parameters are often defined for describing the correlation between $m$ variables \citep{trivedi2007copula}. 

Sklar’s theorem is also valid in the reverse direction, offering a method to generate copula functions.

\begin{equation}
\label{reverse-Sklar’s theorem}
 C_\theta (x_{1},x_{2}, \dots, x_{n})= H({F_{1}}^{-1}(x_{1}),{F_{2}}^{-1}(x_{2}), \dots, {F_{n}}^{-1}(x_{n}))
\end{equation}

In the literature, seven common types of copula functions are used: Gaussian, Clayton, Gumbel, Joe, Ali-Mikhail-Haq (AMH), Frank, and Farlie-Gumbel-Morgenstein (FGM) copulas \citep{trivedi2007copula, nelsen2006introduction}. Among these, Gaussian, Frank, AMH and FGM copulas can capture both positive and negative dependencies, whereas the others measure only positive correlations. Detailed information about the definition and properties of these copula functions is provided in Table \ref{copula functions}. Notably, Gaussian, Frank, and FGM copulas exhibit symmetric tail dependency distributions around the mean, however, FGM generally describes weaker correlations between variables. In contrast, Clayton, Gumbel, AMH and Joe copulas display asymmetric tail dependencies. The Clayton and AMH copula, in particular, shows a more noticeable difference between tails, ensuring high dependency at low values but less correlation at high values, unlike the Gumbel and Joe copulas.

\begin{sidewaystable}[p]
\footnotesize
\centering
\caption{Copula functions}\label{copula functions}
\begin{tabular}{cccc}
\hline
\textbf{$C(F_{1}, F_{2})$} & \textbf{Formulation} & \textbf{Dependence parameter} & \textbf{Kendal's tau} \\
\hline
Gaussian & $\Phi (\Phi^{-1}(F_{1}), \Phi^{-1}(F_{2}); \theta)$ & $ (-1, 1) \setminus \{0\} $ & $\frac{2}{\pi} \sin^{-1} \theta$ \\
\\
Clayton & $(F_{1}^{-\theta} + F_{2}^{-\theta} - 1)^{-\frac{1}{\theta}}$ & $ [-1, \infty) \setminus \{0\} $ & $\frac{\theta}{\theta + 2}$ \\
\\
Gumbel & $\exp \left\{ -(( -\ln F_{1})^\theta + ( -\ln F_{2})^\theta)^{\frac{1}{\theta}} \right\}$ & $ [1, \infty)$ & $\frac{\theta - 1}{\theta}$ \\
\\
Joe & $1 - [(1 - F_{1})^\theta + (1 - F_{2})^\theta - (1 - F_{1})^\theta(1 - F_{2})^\theta]^{\frac{1}{\theta}}$ & $ (1, \infty)$ & $1 + \frac{4}{\theta^2} \int_0^1 t \ln(t) (1 - t)^{\frac{2(1 - \theta)}{\theta}} dt$ \\
\\
AMH & $F_{1}F_{2} (1 - \theta (1 - F_{1})(1 - F_{2}))^-1$ & $[-1, 1] \setminus \{0\}$ & $\frac{3\theta-2}{3\theta}$ - $\frac{2 {(1-\theta)}^2 \ln(1-\theta)}{3{\theta}^2}$\\
\\
Frank & $1 - \frac{1}{\theta} \ln \left( 1 + \frac{(e^{-\theta F_{1}} - 1)(e^{-\theta F_{2}} - 1)}{e^{-\theta} - 1} \right)$ & $ (-\infty, \infty) \setminus \{0\} $ & $1 - \frac{4}{\theta} \left( 1 - \frac{1}{\theta} \int_0^1 \frac{t}{e^t - 1} dt \right)$ \\
\\
FGM & $F_{1}F_{2} (1 + \theta (1 - F_{1})(1 - F_{2}))$ & $[-1, 1] \setminus \{0\}$ & $\frac{2\theta}{9}$\\
\hline
\end{tabular}
\end{sidewaystable}

The work by \citep{bhat2009copula} can be considered as the first effort in transportation literature, utilizing the copula-based approach to jointly model residential neighbourhood choice and daily household vehicle miles of travel (VMT). This study underscores the efficacy of this method in capturing diverse correlation structures among random variables, as opposed to the conventional sample selection approach, which relies on an assumption of bivariate normality on the error terms. Building upon this study, the copula approach is used for several transport areas such as the model of mode choice and escorting pattern selection of students for their school trips \cite{ermagun2015joint}, discrete-continuous model of vehicle type choice and miles of travel \cite{spissu2009copula}, collision type and injury severity of crashes \cite{yasmin2014examining}, mode choice and travel distance \cite{parsa2019does}, the dependence among different traffic variables in the traffic simulation process \cite{fard2019copula}, freight mode choice and shipment size \cite{keya2019joint} and the spatial correlation among close vacant vehicles in car-sharing systems \cite{kostic2021deep}. In the literature, these studies are categorized under the theory-driven approach, where a predetermined functional form is assumed for each dependent variable.

The promising success of machine learning and deep learning algorithms in capturing non-linearity and enhancing model performance has also garnered attention in the field of joint modelling in transportation. In recent years, Copula-Bayesian Network (Copula-BN) is used to investigate the dependency structure between lane-changing (LC) and car-following (CF) maneuvers \cite{chen2021data}, factors influencing the length of railway disruption \cite{zilko2016} and risk factors in the operational metro tunnel \cite{pan2019model}. Furthermore, in other fields of study, for forecasting tasks that involve predicting multiple related time series values, some researchers have integrated sequence modelling tasks like autoregressive models (LSTM-RNN) or the transformer architecture with copula techniques \cite{salinas2019high,drouin2022tactis}. This hybrid approach not only forecasts future values within a deep learning-based sequence-to-sequence model but also effectively captures correlations between interconnected time series factors, resulting in multivariate time series forecasting. These studies highlight recent efforts to leverage the advantages of both data-driven model structures and copula functions in joint modelling. In addition, \citep{golshani2018modeling} generally compared the artificial neural networks model with the copula-based joint model for estimating travel mode and start time. The results of this study show that the neural network model outperforms the copula-based joint model in terms of predictive accuracy. This study seeks to integrate these two modelling frameworks with a distinct objective—to examine the role of deep neural networks in mitigating the effects of unobserved confounders within causal inference frameworks.

\section{METHODOLOGY}
In transportation, researchers highlighted the potential non-causal association between multiple dependent variables, which may result from common unobserved attributes \cite{bhat1997work, brathwaite2018causal}. However, most studies have ignored this association in both association-based and causality-based analysis and researchers barely model the dependent factors, that influence overall decision-making, simultaneously. This study explores the potential of deep learning architectures in isolating unobserved confounding variables for causal analysis. In other words, our proposed model allows us to examine the extent to which deep neural networks can help eliminate non-causal associations—rooted in unobserved factors—from true causal relationships. To this end, we present the frameworks of both the theory-driven and hybrid copula-based joint models, referred to as Copula-Logit and Copula-ResLogit, respectively.

It is important to note that the structure of joint models is generally defined by the number of dependent variables, the number of choices within each dependent variable, and the specific characteristics of the case study. In this study, we present the proposed methodology through two case studies—-pedestrian crossing behaviour and London travel behaviour—-and develop a bivariate Copula-Logit model and a bivariate Copula-ResLogit model for each case study in separate sections. However, the proposed model can be readily generalized to other case studies with different characteristics. For each model structure, various copula functions, along with an independent model that assumes no non-causal unobserved correlation between variables, are evaluated.

\subsection{Copula-ResLogit}
\label{Copula-ResLogit}
The Copula-ResLogit is a hybrid deep copula-based joint model that is primarily based on the theoretical concept of the copula approach combined with the architecture of the ResNet-based models. The Copula-ResLogit model comprises two or multiple model structures, representing dependent variables. The correlation between these variables, stemming from the unobserved variables, is captured using the copula functions, resulting in a dependent multivariate distribution. The ResNet-based components, including ResLogit and Ordinal-ResLogit structures are Residual Neural Network(ResNet)-based Random Utility Maximization (RUM) models designed for both categorical and ordinal datasets \cite{wong2021reslogit,kamal2024ordinal}. The ResNet blocks in Copula-ResLogit try to account for the effect of confounders or shared unobserved variables. In fact, not only does our approach aim to eliminate unobserved confounding impacts through the ResNet blocks, but also uncovers any remaining unobserved confounding effects in the causal analysis. The structure of the Copula-ResLogit model is illustrated in Figure \ref{Copula-ResLogit}. In this figure, the human brain symbolizes the decision-making process, where multiple factors may influence each other, and an individual's overall decision is made by considering the association between these correlated factors that can be causal or confounding associations. In causal analysis and what-if (counterfactual) scenarios, the focus should be on isolating direct causal relationships. Accordingly, this study investigates the capability of deep ResNet blocks to eliminate non-causal associations rooted in unobserved confounders.

\begin{figure}[!ht]
  \begin{center}
   \includegraphics[width=1 \textwidth]{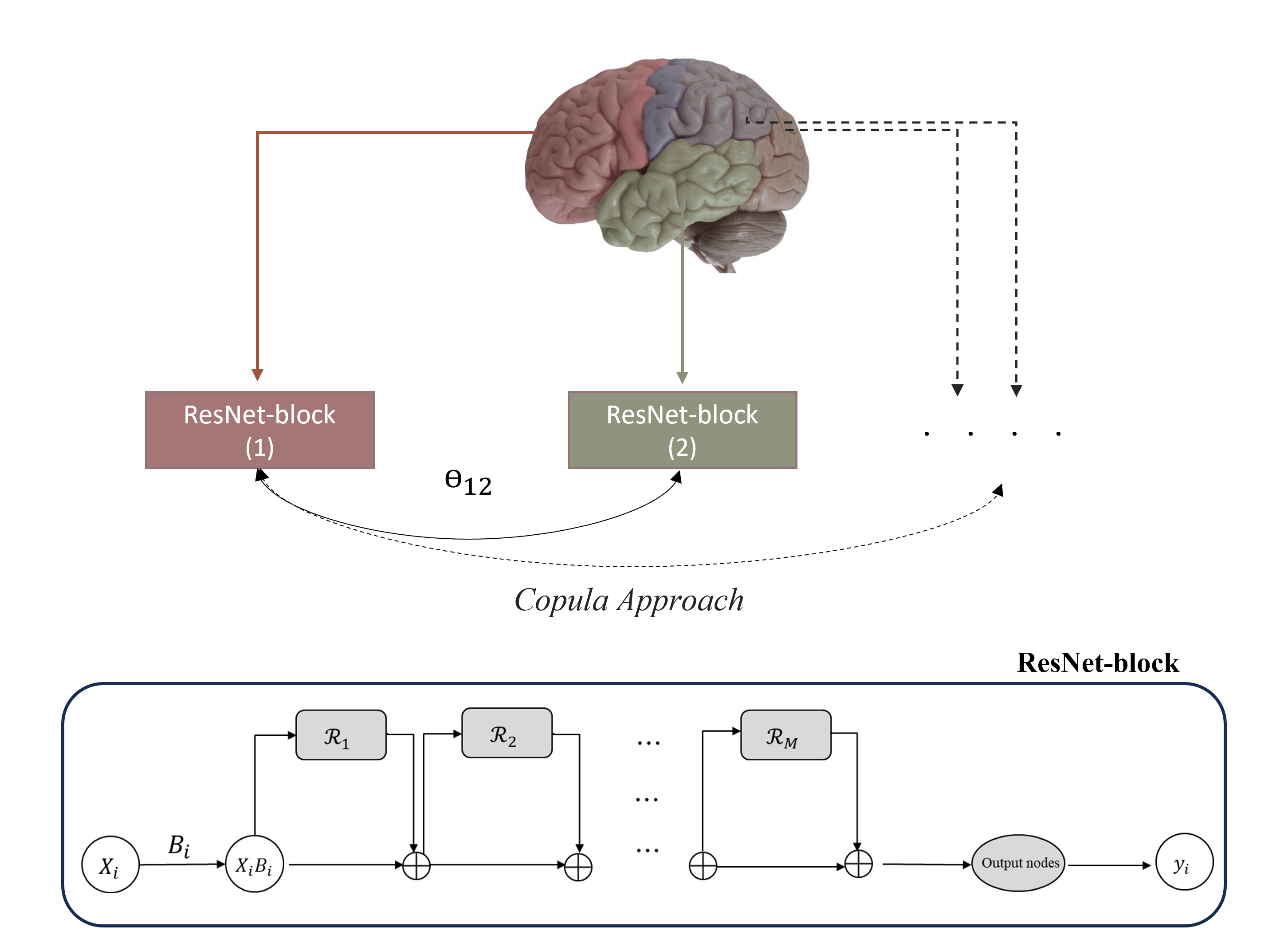}
   \caption{Copula-ResLogit Model Structure}\label{Copula-ResLogit}
   \end{center}
\end{figure}

The Copula-ResLogit model can be categorized as an extension of the ResNet-based model structure, primarily focusing on unobserved confounding association in causal analysis and joint modelling. In the following sections the model specification of Copula-ResLogit is presented for the ordinal-ordinal model of stress and wait-time pedestrian behaviour and the multinomial-ordinal model of travel mode and travel distance studies.

\subsubsection{Copula-ResLogit for ordinal-ordinal joint modelling}
\label{Pedestrian crossing behaviour}
We aim to examine the potential non-causal association between two key factors in pedestrian crossing behaviour: pedestrian stress level and waiting time, after including deep ResNet blocks. In this study, we investigate the ordinal forms of waiting time and stress levels, which provide a more informative and realistic representation of real-world comprehension compared to a quantitative scale. In fact, in the real world, many responses like an individual's socioeconomic status are observed not on a continuous scale, but on an ordinal scale and often described using ordinal levels such as low, medium, and high. Therefore, to jointly model the ordered levels of pedestrians' stress levels and their waiting time, we utilize Ordinal-ResLogit frameworks from ResNet-based model structures \cite{kamal2024ordinal}. It is worth noting that in cases involving categorical dependent variables, they can be modelled using the ResLogit framework instead.

For each pedestrian $q$, let $i$ and $k$ be an index representing the stress level and waiting time, respectively. Building upon the Ordinal-ResLogit framework, the utility function for ordinal categories is formulated to account for both observed variables ($x_{qi}$ and $z_{qk}$), as well as unobserved factors, which is captured through the residual deep component denoted as $(g_{q})$. In addition, the Ordinal-ResLogit framework reduces the continuous problem to multiple binary classifiers associated to ordinal categories \cite{kamal2024ordinal,cao2020rank}. This framework ensures rank monotonicity between ordered predictions by applying the deep CORAL Layer without the need for a cost function \citep{cao2020rank}. Therefore, after capturing the unobserved impacts through residual layers, the corresponding output is fed into the CORAL layer, reducing the problem to $K-1$ binary classifiers.

\begin{equation}\label{stressUtility-Copula-ResLogit}
s^{*}_{qi} = \sum_{i=1}^{I} w_i (\beta_{i} x_{qi}  + g_{qi}) + b_i + \varepsilon_{qi} = v^{s}_{qi} + \varepsilon_{qi}
\end{equation}

\begin{equation}\label{waitUtility-Copula-ResLogit}
w^{*}_{qk} = \sum_{k=1}^{K} w_k (\gamma_{k} z_{qk} + g_{qk}) + b_k + \eta_{qk} =v^{w}_{qk} + \eta_{qk} 
\end{equation}

\noindent where $w_k$ and $b_k$ are parameters of the deep CORAL Layer of Ordinal-ResLogit in which the problem is converted to $k-1$ binary classification problem. Then the probability of each binary classifier for waiting time and stress level structures can be computed based on the associated propensity $(s*_{qi}$, $w*_{qk})$ and the marginal distributions $F_{\varepsilon}(.)$ and $F_{\eta}(.)$ as follows:

\begin{equation}
\label{stressBinaryprediction}
Pr(s_{qi} = 1) = F_{\varepsilon}(v^{s}_{qi}) = \sigma (\sum_{i=1}^{I} w_i (\beta_{i} x_{qi}  + g_{qi}) + b_i)
\end{equation}

\begin{equation}
\label{waitBinaryprediction}
Pr(w_{qk} = 1) = F_{\eta}(v^{w}_{qk}) =\sigma (\sum_{k=1}^{K} w_k (\gamma_{k} z_{qk} + g_{qk}) + b_k) 
\end{equation}

Detailed information on the Ordinal-ResLogit framework can be found in our previous study \cite{kamal2024ordinal}. Accordingly, the probability of transiting from a lower level to a higher level is calculated as follows:

\begin{equation} 
\label{stresslevelprediction-Copula-ResLogit}
Pr(s_{q}=i)  = Pr(s_{qi-1} = 1) - Pr(s_{qi} = 1)
\end{equation}

\begin{equation} 
\label{waitlevelprediction-Copula-ResLogit}
Pr(w_{q}=k) =  Pr(w_{qk-1} = 1) - Pr(w_{qk} = 1)
\end{equation}

In our study, $i \in \{low, high\}$ and $k \in \{low, high\}$. It is of note that for ordinal responses, the lowest category is always selected initially, and we are primarily interested in whether individuals choose the high level or not. Therefore, the joint distribution of four potential conditions, where an individual $q$ experiences stress level $i$ and selects waiting time $k$ is determined by the following equations:

\begin{equation}\label{JointDist of Copula-ResLogit1}
\begin{split}
Pr \left( s_q = 1, \ w_q = 1 \right) & =Pr \left( s^{*}_{q}  < 0, \ w^{*}_{q}  < 0\right) = Pr (\varepsilon_{q}  < -v^{s}_{qi}, \eta_{q}  < -v^{w}_{qk} )
\end{split}
\end{equation}

\begin{linenomath}
  \begin{equation}\label{JointDist of Copula-ResLogit2}
  \begin{split}
    Pr \left( s_q = 1, \ w_q = 2 \right) & =Pr \left( s^{*}_{q}  < 0, \ w^{*}_{q}  > 0\right) = Pr (\varepsilon_{q}  < -v^{s}_{qi}, \eta_{q}  > -v^{w}_{qk} )
  \end{split}
  \end{equation}
\end{linenomath}

\begin{linenomath}
  \begin{equation}\label{JointDist of Copula-ResLogit3}
  \begin{split}
    Pr \left( s_q = 2, \ w_q = 1 \right) & =Pr \left( s^{*}_{q}  > 0, \ w^{*}_{q}  < 0\right) = Pr (\varepsilon_{q}  > -v^{s}_{qi}, \eta_{q}  < -v^{w}_{qk} )
  \end{split}
  \end{equation}
\end{linenomath}

\begin{linenomath}
  \begin{equation}\label{JointDist of Copula-ResLogit4}
  \begin{split}
    Pr \left( s_q = 2, \ w_q = 2 \right) & =Pr \left( s^{*}_{q}  > 0, \ w^{*}_{q}  > 0\right) = Pr (\varepsilon_{q}  > -v^{s}_{qi}, \eta_{q}  > -v^{w}_{qk} )
  \end{split}
  \end{equation}
\end{linenomath}

The calculation of the above probability function depends on the joint distribution of the two random variables which based on the Equations \ref{Sklar’s theorem} and \ref{Joint dist} can be replaced by the copula function as follows:

\begin{linenomath}
  \begin{equation}\label{Copula-ResLogit1}
  \begin{split}
    Pr \left( s_q = 1, \ w_q = 1 \right) & = C_\theta \left(F_{\varepsilon}(-v^{s}_{qi}) , \ F_{\eta} (-v^{w}_{qk})\right)  \end{split}
  \end{equation}
\end{linenomath}

\begin{linenomath}
  \begin{equation}\label{Copula-ResLogit2}
  \begin{split}
    Pr \left( s_q = 1, \ w_q = 2 \right) & = F_{\varepsilon}(-v^{s}_{qi}) - C_\theta \left( F_{\varepsilon}(-v^{s}_{qi}) , \ F_{\eta} (-v^{w}_{qk})\right)
  \end{split}
  \end{equation}
\end{linenomath}

\begin{linenomath}
  \begin{equation}\label{Copula-ResLogit3}
  \begin{split}
    Pr \left( s_q = 2, \ w_q = 1 \right) & = \ F_{\eta} (-v^{w}_{qk}) - C_\theta \left(F_{\varepsilon} (-v^{s}_{qi}) , \ F_{\eta} (-v^{w}_{qk})\right)
  \end{split}
  \end{equation}
\end{linenomath}

\begin{linenomath}
  \begin{equation}\label{Copula-ResLogit3}
  \begin{split}
    Pr \left( s_q = 2, \ w_q = 2 \right) & =  F_{\varepsilon}(v^{s}_{qi}) -\ F_{\eta} (-v^{w}_{qk}) + C_\theta \left(F_{\varepsilon}(-v^{s}_{qi}) , \ F_{\eta} (-v^{w}_{qk} \right)
  \end{split}
  \end{equation}
\end{linenomath}

\subsubsection{Copula-ResLogit for multinomial-ordinal joint modelling}
\label{Pedestrian crossing behaviour}
For this case study, we examines to what extend the deep ResNet structure is able to block the potential non-causal association between travel mode and distance choices. To jointly model the multinomial travel mode choice and ordered levels of travel distance for non-mandatory trips, we utilize ResLogit and Ordinal-ResLogit frameworks from ResNet-based model structures, respectively \cite{wong2021reslogit, kamal2024ordinal}. Detailed information on ResLogit model structure can be found in \citep{wong2021reslogit,kamal2024ordinal}.

For each individual $q$, let $j$ and $r$ be an index representing the travel mode choice and level of travel distance, respectively. Building upon the ResLogit and Ordinal-ResLogit framework, the utility functions, accounting for both observed variables and unobserved heterogeneity, are defined as follows:

\begin{equation}\label{modechoiceUtility-Copula-ResLogit}
m^{*}_{qj} = \beta_{j} x_{qj}  + g_{qj} + \varepsilon_{qj} = v^{m}_{qj} + \varepsilon_{qj}
\end{equation}

\begin{equation}\label{traveldistanceUtility-Copula-ResLogit}
d^{*}_{qr} = \sum_{r=1}^{R} w_r (\gamma_{r} z_{qr} + g_{qr}) + b_r + \eta_{qk} =v^{d}_{qr} + \eta_{qr} 
\end{equation}

In our study, $j \in \{active modes, private car, public transit\}$ and $r \in \{short, long\}$. According to utility maximization theory, a rational individual $q$ would choose mode $j$ associated with the highest utility. In other words, individual $q$ chooses mode $j$ among all modes if:

\begin{equation}
m^{*}_{qj} > Max_{i \neq j} (m^{*}_{qi})
\end{equation}

\begin{equation}
\tau_{qj} = Max_{i \neq j}  (m^{*}_{qi}) - \varepsilon_{qj}
\end{equation}
 
The joint distribution of six potential choices, where an individual $q$ selects travel mode $j$ and travel distance $r$, is determined by the following equations:

\begin{equation}
\begin{split}
Pr \left( m_q = j, \ d_q = 1 \right) & = Pr \left( m^{*}_{qj} > Max_{i \neq j} (m^{*}_{qi}), \ d^{*}_{q}  < 0\right) = Pr \left( \tau_{qj}  < v^{m}_{qj}, \eta_{q}  < -v^{d}_{q} \right)
\end{split}
\end{equation}

\begin{equation}
\begin{split}
Pr \left( m_q = j, \ d_q = 2 \right) & = Pr \left( m^{*}_{qj} > Max_{i \neq j} (m^{*}_{qi}), \ d^{*}_{q}  > 0\right) = Pr \left( \tau_{qj}  < v^{m}_{qj}, \eta_{q}  > -v^{d}_{q} \right)
\end{split}
\end{equation}

The calculation of the above probability function depends on the joint distribution of the two random variables. Based on the Equations \ref{Sklarstheorem} and \ref{Joint dist} this function can be replaced by the copula function as follows:

\begin{linenomath}
  \begin{equation}\label{Copula-ResLogit3}
  \begin{split}
    Pr \left( m_q = j, \ d_q = 1 \right)  & = C_\theta \left(F_{\varepsilon} (v^{m}_{qj}) , \ F_{\eta} (-v^{d}_{q})\right)
  \end{split}
  \end{equation}
\end{linenomath}

\begin{linenomath}
  \begin{equation}\label{Copula-ResLogit3}
  \begin{split}
    Pr \left( m_q = j, \ d_q = 2 \right)  & = F_{\varepsilon}(v^{m}_{qj}) - C_\theta \left(F_{\varepsilon} (v^{m}_{qj}) , \ F_{\eta} (-v^{d}_{q})\right)
  \end{split}
  \end{equation}
\end{linenomath}

\subsubsection{Model estimation}
In order to estimate the parameters of Copula-ResLogit, denoted by $\lambda$, including $\beta$,  $\gamma$, residual weights, CORAL parameters and the copula parameter $\theta$, the cost function of stress-wait time and mode-distance problems are calculated as follows: 

\begin{equation}
\label{loss function-Copula-ResLogit}
LL(\lambda)  = \sum_{q=1}^{Q}\sum_{i=1}^{I} \sum_{k=1}^{K} \Bigg[ln\ Pr(s_q = i, \ w_q = k)I[s_{qi} = 1]  I[w_{qk} = 1])\Bigg]
\end{equation}

\begin{equation}
\label{loss function-Copula-ResLogit}
LL(\lambda)  = \sum_{q=1}^{Q}\sum_{j=1}^{J} \sum_{r=1}^{R} \Bigg[ln\ Pr(m_q = j, \ d_q = r )I[m_{qj} = 1]  I[d_{qr} = 1])\Bigg]
\end{equation}

\noindent where generally,

\begin{equation}
    I[x = 1] =
    \begin{cases}
        1, &\text{if }  x = 1 \\
        0, &\text{if }  x = 0
    \end{cases}
\end{equation}

It is of note that in our Ordinal-ResLogit model structure, we assume that all binary classifiers have the same importance in optimization. For estimation, we use the mini-batch data-driven stochastic gradient descent-based (SGD) learning algorithm and apply an RMSprop optimization step. In addition, an early-stopping approach is applied to stop the training process once the performance of the model on the validation dataset starts to degrade \cite{goodfellow2016deep}. We implement the Copula-ResLogit using open-source deep learning libraries Python. 

\subsubsection{Hyperparameters}
The performance of Copula-ResLogit considerably depends on model hyperparameter settings. The hyperparameters of our model are the number of residual layers, learning rate and batch size. In the process of modelling, we divided the dataset into two subsets using a 70:30 training/validation split. In this study, we employ the random search technique to optimize the hyperparameters of our model. This process entails specifying a set of values for each hyperparameter and randomly selecting combinations from these sets to train the model on the training dataset. In addition, for the learning algorithm, an RMSprop optimization step is applied to scale the learning rate based on model parameters. Our previous studies showed that increasing the depth of the ResNet-based models guarantees better performance \cite{wong2021reslogit, kamal2024ordinal}. However, to control the complexity of our model and time computation, we eventually opted for a configuration consisting of 16 residual layers. In addition, a batch size of 64 is chosen for training the model using the mini-batch SGD learning algorithm. 

\subsubsection{Model validation}
To evaluate and compare the Copula-ResLogit models developed using different copula functions, the Akaike Information Criterion (AIC) and and Mean Prediction Error (MPE) on out-of-sample (unseen sample) are used. AIC considers the complexity of model and is calculated by:

\begin{equation}
\label{AIC}
AIC  = 2LL(\lambda) + 2B
\end{equation}

\noindent where $B$ is the number of estimated parameters. The Akaike information criterion (AIC) is based on information theory. Therefore, a model with a lower AIC or less lost information shows relatively better performance. AIC balances goodness of fit with model complexity, but with a lighter penalty on the number of parameters compared to Bayesian Information Criterion (BIC). The considerable difference in the number of parameters between deep learning-based models and traditional models notably influences the complexity component in both AIC and BIC approaches. To achieve a better balance between goodness-of-fit and model complexity, AIC is chosen for model evaluation, as BIC assigns greater weight to the complexity component. 

The Mean Prediction Error (MPE) provides information about the accuracy of the models on the validation dataset and is defined as follows:

\begin{equation}
\label{MPE}
MPE  = \frac{1}{Q} \sum_{q=1}^{Q} I[y_{q} \neq \hat{y}_{q}]
\end{equation}

\noindent 
where $y_{q}$ and ${\hat{y}}_{q}$ represent the actual and predicted joint choices for the combination of stress level $i$ and waiting time $k$.

\subsection{Copula-Logit}
To examine the power of the deep learning component in controlling for the effects of unobserved confounders, a completely theory-driven copula-based joint model is developed, called copula-Logit. In other words, the copula-Logit framework does not block unobserved confounding impacts. In Copula-Logit, the Proportional Odds Model (POM) or Ordered Logit Model is employed to model the ordinal responses and the Multinomial Logit Model (MNL) is applied in cases involving categorical dependent variables. The ordered logit model associates a single utility function, known as a latent propensity, to ordinal responses using estimated thresholds \cite{mccullagh1980regression}.

\subsubsection{Copula-Logit for ordinal-ordinal joint modelling}
Based on the ordered logit model, for ordinal responses of pedestrian stress level and wait time, the latent stress $(s^{*})$ and waiting time $(w^{*})$ is specified as a linear function of explanatory variables $x_{q}$ and $z_{q}$ with associated parameters $\beta$ and $\gamma$ which remain constant between ordinal categories as follows:

\begin{equation}
\label{stressUtility-Copula-Logit}
S^{*}_{q}=\beta x_{q} + \varepsilon_{q} 
\end{equation}

\begin{equation}
\label{waitUtility-Copula-Logit}
W^{*}_{q} = \gamma z_{q} + \eta_{q}   
\end{equation}

In these equations, $\varepsilon$ and $\eta$ are random error terms assumed to be logistic. These latent propensities relate to the ordinal responses through the thresholds as follows:

\begin{equation} 
s_{q} = i \ \text{if} \ \delta_{i-1} < s^{*}_{q} \leq \delta_i
\end{equation}

\begin{equation} 
w_{q} = k \ \text{if} \ \psi_{k-1} < w^{*}_{q} \leq \psi_k
\end{equation}

\noindent where the thresholds must satisfy the constraint $\delta_1 \leq \delta_2 \leq ... \leq \delta_i$ and  $\psi_1 \leq \psi_2 \leq ... \leq \psi_k$. Typically, it is assumed that $\delta_0=- \infty$ , $\delta_i= + \infty$ and $\psi_0=- \infty$ , $\psi_k= + \infty$.
 
For each crossing component, the independent probability relies on the marginal distribution of $\varepsilon$ and $\eta$ and is calculated based on the cumulative probabilities of the response variable. The joint distribution of experiencing stress level $i$ and choosing waiting time $k$ is determined by Equation \ref{JointDist of Copula-Logit}.

\begin{linenomath}
  \begin{equation}\label{JointDist of Copula-Logit}
  \begin{split}
    Pr \left( s_q = i, \ w_q = k \right) & =Pr \left(  \delta_{i-1} < s*_q  < \delta_i, \ \psi_{k-1} < w*_q  < \psi_k \right) \\
    & = Pr \left(  \delta_{i}-\beta x_{q} < \varepsilon_{q}  < \delta_{i-1}-\beta x_{q}, \ \psi_{k}-\gamma z_{q} < \eta_{q}  < \psi_{k-1}-\gamma z_{q} \right) \\
    & = Pr \left( \varepsilon_{q} < \delta_{i}-\beta x_{q} , \eta_{q} < \psi_{k}-\gamma z_{q} \right) \\
    & - Pr \left( \varepsilon_{q} < \delta_{i}-\beta x_{q} , \eta_{q} < \psi_{k-1}-\gamma z_{q} \right) \\
    & -Pr \left( \varepsilon_{q} < \delta_{i-1}-\beta x_{q} , \eta_{q} < \psi_{k}-\gamma z_{q} \right) \\
    & + Pr \left( \varepsilon_{q} < \delta_{i-1}-\beta x_{q} , \eta_{q} < \psi_{k-1}-\gamma z_{q} \right)
  \end{split}
  \end{equation}
\end{linenomath}

Similar to Copula-ResLogit model, these joint probability functions rely on the joint distribution of the two random variables. Using Equations \ref{Sklarstheorem} and \ref{Joint dist}, the copula function can be substituted into Equation \ref{JointDist of Copula-Logit}.

\begin{linenomath}
  \begin{equation}\label{Copula-Logit}
  \begin{split}
    Pr \left( s_q = i, \ w_q = k \right) & = C_\theta \left(\left( F_{\varepsilon}(\delta_{i}-\beta x_{q} \right) , \ F_{\eta} \left(\psi_{k}-\gamma z_{q}\right)\right) \\
    & - C_\theta \left(\left( F_{\varepsilon}(\delta_{i}-\beta x_{q} \right) , \ F_{\eta} \left(\psi_{k-1}-\gamma z_{q}\right)\right)\\
    & - C_\theta \left(\left( F_{\varepsilon}(\delta_{i-1}-\beta x_{q} \right) , \ F_{\eta} \left(\psi_{k}-\gamma z_{q}\right)\right) \\
    & + C_\theta \left(\left( F_{\varepsilon}(\delta_{i-1}-\beta x_{q} \right) , \ F_{\eta} \left(\psi_{k-1}-\gamma z_{q}\right)\right)
  \end{split}
  \end{equation}
\end{linenomath}

\noindent where $F_{\epsilon}(.)$ and $F_{\eta}(.)$ are the marginal distributions following the logistic distribution. 

\subsubsection{Copula-Logit for multinomial-ordinal joint modelling}
In the case of multinomial travel mode choice and the ordered levels of travel distance for non-mandatory trips, the multinomial logit model and ordered logit model are used and the utility functions are defined as follows:

\begin{equation}\label{modechoiceUtility-Copula-ResLogit}
m^{*}_{qj} = \beta_{j} x_{qj}  + \varepsilon_{qj} 
\end{equation}

\begin{equation}
\label{waitUtility-Copula-Logit}
d^{*}_{q} = \gamma z_{q} + \eta_{q}   
\end{equation}

\begin{equation} 
d_{q} = r \ \text{if} \ \psi_{r-1} < d^{*}_{q} \leq \psi_r
\end{equation}

The joint distribution of six potential choices, where an individual $q$ selects travel mode $j$ and travel distance $r$, is determined by the following equations:

\begin{linenomath}
  \begin{equation}\label{Copula-Logit}
  \begin{split}
    Pr \left( m_q = j, \ d_q = 1 \right) & =   Pr \left( \tau_{qj}  < v^{m}_{qj}, \psi_{r-1} < d^{*}_{q} \leq \psi_r \right) \\
    & = Pr \left(  \tau_{qj}  < v^{m}_{qj}, \eta_{q} < \psi_{r}-\gamma z_{q} \right)- Pr \left(  \tau_{qj}  < v^{m}_{qj}, \eta_{q} < \psi_{r-1}-\gamma z_{q} \right)
  \end{split}
  \end{equation}
\end{linenomath}

This equation can be replaced by the copula function as follows: 

\begin{linenomath}
  \begin{equation}\label{Copula-ResLogit3}
  \begin{split}
    Pr \left( m_q = j, \ d_q = 1 \right)  & = C_\theta \left(F_{\varepsilon} (\beta_{j} x_{qj}) , \ F_{\eta} (\psi_{r}-\gamma z_{q})\right) - C_\theta \left(F_{\varepsilon} (\beta_{j} x_{qj}) , \ F_{\eta} (\psi_{r-1}-\gamma z_{q})\right)
  \end{split}
  \end{equation}
\end{linenomath}

Similar to Copula-ResLogit, the parameters of Copula-Logit are estimated using the maximum likelihood function. In addition, developed models are evaluated based on Akaike Information Criterion (AIC) and Mean Prediction Error (MPE).

\section{Case Study}
For the case of pedestrian crossing behaviour, a Virtual Reality (VR) dataset collected based on the Virtual Immersive Reality Experiment (VIRE) \cite{farooq2018virtual} is used. The dataset consists of 1,406 pedestrian responses in the presence of Automated Vehicles (AVs) on the sidewalk \cite{mudassar2021analysis}. In addition to controlled factors including street characteristics, traffic conditions, environmental situations and pedestrians' socio-demographic and travel patterns, Galvanic Skin Response (GSR) sensors were utilized during the experiments to measure the relative stress levels of the participants \cite{mudassar2021analysis}. The description of explanatory variables used in this study is shown in Table \ref{pedestriandataset}. This sensor operates by applying a small electrical charge to measure sweat production on an individual’s finger, with higher charges indicating greater sweat production and, consequently, higher stress levels. In our study, the mean stress level for each pedestrian scenario was normalized based on the minimum and maximum stress levels observed within that scenario.

\begin{table}[p]
\footnotesize
    \caption{Description of explanatory variables of pedestrian crossing behaviour} \label{pedestriandataset}
	\begin{center}
		\begin{tabular}{l p{9cm} l >{\centering\arraybackslash}p{1.5cm} c l}
		    \\\hline
			Variable & Description & Mean & Standard deviation \\\hline
			\emph{\textbf{Road attributes}} \\
			\small{Low lane width}   & \small{1: If the lane width is less than 2.75 meter, 0:otherwise} & 0.334  & 0.472\\
			\small{High lane width} & \small{1: If the lane width is greater than 2.75 meter, 0:otherwise} & 0.359 & 0.480\\
			\small{Two way with a median} & \small{1: If the road type is two way with median, 0:otherwise} & 0.338 & 0.473 \\
			\small{Two way}   & \small{1: If the road type is two way, 0:otherwise} & 0.317 & 0.466 \\
			\small{One way} & \small{1: If the road type is one way, 0:otherwise} & 0.344 & 0.475\\
                \small{Traffic Density} & \small{Density of road (veh/hr/ln)}  & 19.835 & 7.158\\\hline
			\emph{\textbf{Traffic condition}} \\
			\small{Mixed traffic condition}   & \small{1: If traffic in scenario consists of automated vehicles and human-driven vehicles, 0:otherwise} & 0.045 & 0.207 \\
			\small{Fully automated condition}  & \small{1: If traffic in scenario consists of only automated vehicles, 0:otherwise} & 0.925 & 0.263 \\
			\small{Human-driven condition}   & \small{1: If traffic in scenario consists of only human-driven vehicles, 0:otherwise} & 0.029 & 0.170\\ \hline
   		\emph{\textbf{Crossing Attributes}} \\
			\small{Wait Time} & \small{High: If participant’s Wait Time on the sidewalk is greater than 5 seconds, Low: otherwise} & 0.357 & 0.479 \\
			\small{Stress Level}  & \small{High: If participant’s normalized stress is greater than 0.59, Low: otherwise} & 0.252 & 0.435 \\ \hline
			\emph{\textbf{Socio-demographic}} \\
			\small{Age 18-30} & \small{1: If participant’ age is between 18 and 30, 0:otherwise} & 0.542  & 0.498 \\
			\small{Age 30-39} & \small{1: If participant’ age is between 30 and 39, 0:otherwise} & 0.339  & 0.474 \\
			\small{Age 40-49} & \small{1: If participant’ age is between 40 and 49, 0:otherwise} & 0.034  & 0.182 \\
			\small{Age over 50} & \small{1: If participant’ age is more than 50, 0:otherwise} & 0.084  & 0.278 \\
			\small{Female} & \small{1: If participant is female, 0:otherwise} & 0.382  & 0.486\\
			\small{Driving license} & \small{1: If participant has a driving license, 0:otherwise} & 0.895  & 0.307\\
			\small{No car} & \small{1: If participant has no car in the household, 0:otherwise} & 0.302  & 0.345\\
			\small{One car }& \small{1: If participant has one car in the household, 0:otherwise} & 0.345  & 0.476\\
			\small{Over one car} & \small{1: If participant has more than one car in the household, 0:otherwise} & 0.353  & 0.478\\
			\small{Active mode} & \small{1: If participant use active modes regularly, 0:otherwise} & 0.258  & 0.438\\
			\small{Private car mode} & \small{1:If participant use private car regularly, 0:otherwise} & 0.332  & 0.471\\
			\small{Public mode} & \small{1: If participant use transit regularly, 0:otherwise} & 0.409  & 0.492\\
			\small{VR experience} & \small{1: If participant has VR experience, 0:otherwise} & 0.425  & 0.495\\ \hline
			\emph{\textbf{Environment condition }} \\
			\small{Night} & 1\small{: If the time of scenario is night, 0:otherwise} & 0.371  & 0.483\\
			\small{Snowy} & \small{1: If the weather of scenario is snowy, 0:otherwise} & 0.293  & 0.455\\
                \hline
			\hline
		\end{tabular}
	\end{center}
\end{table}

Moreover, we use an openly available real-world dataset, the London Travel Demand Survey (LTDS) \cite{hillel2018recreating} for the case of multinomial-ordinal joint modelling. This dataset consists of individual trip records from April 2012 to March 2015 with corresponding trip trajectories and mode alternatives obtained from a directions Application Programming Interface (API). The present study focuses only on non-mandatory trips—those for which individuals have flexibility in choosing both their travel mode and destination. Table \ref{londondata} summarizes the explanatory variables incorporated in the analysis.

\begin{table}[!ht]
    \caption{Description of explanatory variables of London travel behaviour dataset}\label{londondata} 
    \begin{center}
    \begin{adjustbox}{width=\textwidth}
        \begin{tabular}{l p{9.5cm} l >{\centering\arraybackslash}p{2cm} c l}
            \\\hline
            Variable & Description & Mean & Standard deviation \\\hline
            \emph{\textbf{Travel Attributes}} \\
            
            Travel Mode & & & \\
            \multicolumn{1}{c}{\small{Private car}}   & \small{1: if individual uses private car, 0: otherwise} &   0.478 & 0.450\\
            \multicolumn{1}{c}{\small{Public transit}} & \small{1: if individual uses public transportation, 0: otherwise } & 0.296 & 0.456\\
            \multicolumn{1}{c}{\small{Active modes}} & \small{1: if individual uses bike or walking, 0: otherwise } & 0.226 & 0.456\\
            Travel Cost & & & \\
            \multicolumn{1}{c}{\small{Transit cost}} & \small{Travel cost for transit mode (£)} & 1.456 & 1.287 \\
            \multicolumn{1}{c}{\small{Driving cost}}   & \small{Total travel cost for private car mode (£)} & 1.423 & 2.853 \\
            Travel Distance & \small{Long: more than 7.8 kilometers away from their starting point, Medium: otherwise} & 0.529 & 0.499 \\
            \hline
            
            \emph{\textbf{Socio-demographic}} \\
            \small{Age Group} & & & \\
            \multicolumn{1}{c}{\small{ 18-30}} & \small{1: if individual’s age is between 18 and 30, 0: otherwise} &0.185 & 0.388 \\
            \multicolumn{1}{c}{\small{ 30-45}} & \small{1: if individual’s age is between 30 and 45, 0: otherwise} & 0.315  & 0.465 \\
            \multicolumn{1}{c}{\small{45-60}} & \small{1: if individual’s age is between 45 and 60, 0: otherwise} & 0.232  & 0.422 \\
            \multicolumn{1}{c}{\small{ Over 60}} & \small{1: if individual’s age is more than 60, 0: otherwise} & 0.268  & 0.443 \\
            \small{Gender} & \small{1: if individual is female, 0: otherwise} & 0.557  & 0.497\\
            \small{Disability} & \small{1: If participant has a driving license, 0:otherwise} & 0.895  & 0.307\\
            \small{Driving License} & \small{1: if individual has a driving license, 0: otherwise} & 0.718  & 0.450\\
            \small{Unrestricted Car Access} & \small{1: if there is one or more than one car in the household, 0: otherwise} & 0.271  & 0.444\\
            \hline
        \end{tabular}
    \end{adjustbox}
    \end{center}
    \label{Table6}
\end{table}

\begin{figure}
    \centering
    \subfloat{{\includegraphics[scale=.6]{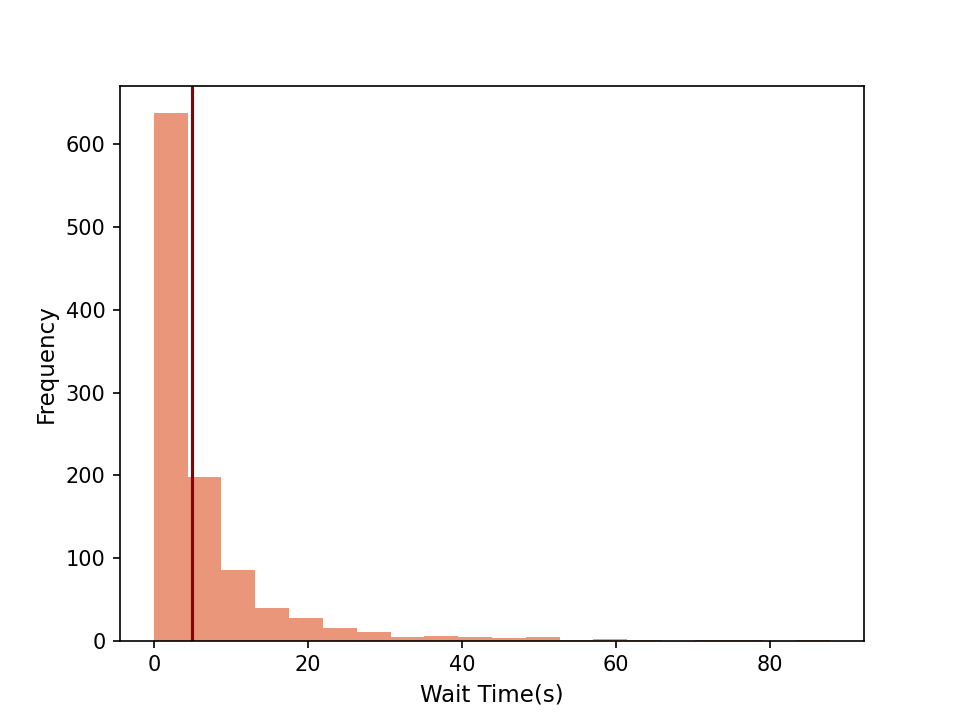}}}
    \subfloat{{\includegraphics[scale=.6]{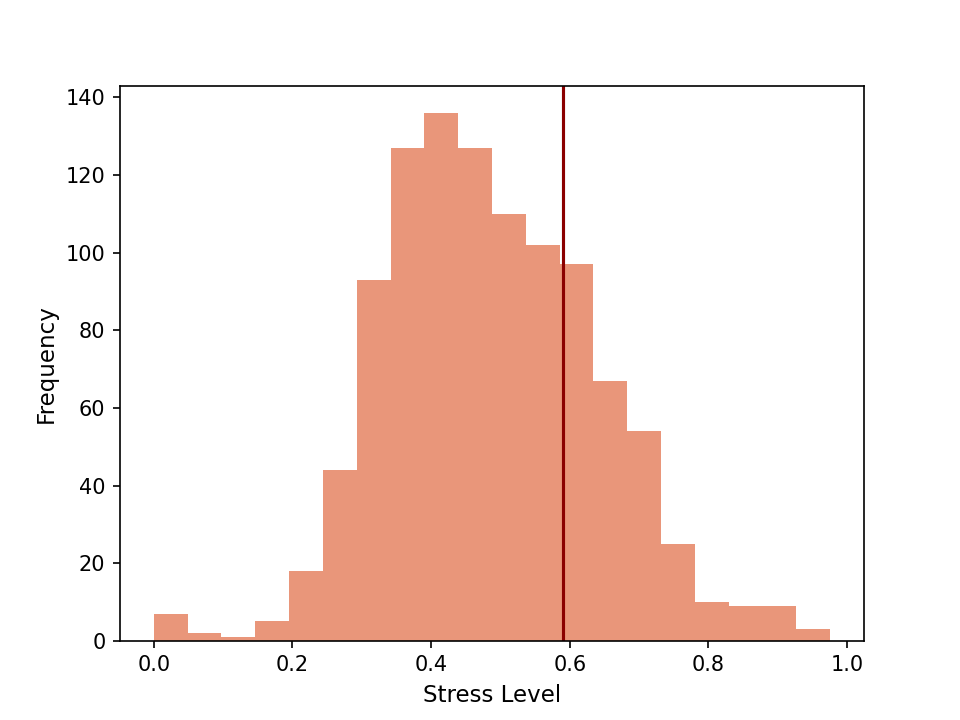}}}
    \caption{Wait time and stress level distribution}
    \label{data distribution}
\end{figure}

It is worth mentioning that in this study, pedestrian wait time, stress level and travel distance variables are discretized into discrete categories using Jenks Natural Breaks classification method and logical facts \cite{jiang2013head}. The chosen method functions similarly to the K-means algorithm but is applied in a univariate context. It classifies a given data into several groups in such a way as to minimize the variance of members of each group while maximizing the variance between different groups. The distributions of normalized stress level and waiting time are shown in Figure \ref{data distribution}. In addition, the optimum thresholds of different categories for these variables are presented in Table \ref{pedestriandataset} and \ref{londondata}.

\section{Results and Discussion}
In this study, we develop both Copula-ResLogit and Copula-Logit models for each case study to assess the potential non-causal dependencies between dependent variables and to evaluate the capacity of deep learning techniques to mitigate confounding effects. These models are built using four flexible copula functions—Frank, FGM, Gaussian, and AMH—which are capable of capturing a wide range of dependency structures, including both positive and negative correlations. In addition, the independent model structure is developed to compare against the joint model structure, enabling us to assess the idea of dealing with unobserved confounders using deep learning blocks in model structures. Section \ref{Results-pedestrian} and \ref{Results-london} are compared the performance of Copula-ResLogit and Copula-Logit. In addition, a brief discussion is provided on the estimated parameters of the significant observed variables. In each section, the performance of all developed models is assessed based on their copula properties, Akaike Information Criterion (AIC) and Mean Prediction Error (MPE).

\subsection{Pedestrian stress and waiting time}
\label{Results-pedestrian}
The performance of all developed models for the pedestrian crossing dataset is shown in Table \ref{ModelPerformanceResults-pedestrian}. The copula parameters are estimated within the acceptable range for all copulas. The results indicate that in the Copula-ResLogit and Copula-Logit models, the independent structure (Product-ResLogit) and the Ali-Mikhail-Haq (AMH) function (AMH-Logit), with a lower AIC perform better than the other copula functions, respectively. In addition, the Product-ResLogit model outperforms the AMH-Logit model. In fact, although the number of parameters in the residual deep layer, which represents the complexity of the deep-based copula model, is treated as a penalty in the model selection method, the lower AIC and higher accuracy of the Product-ResLogit model underscore the significance of deep components in enhancing model performance compared to AMH-Logit.

\begin{table}[!htbp]
\footnotesize
\centering
\caption{Model Performance of Copula-ResLogit and Copula-Logit for pedestrian stress and waiting time}\label{ModelPerformanceResults-pedestrian}
\begin{tblr}{
  colspec = {lcccccccc},
  cell{1}{2,6} = {c=2}{c}, 
}
\hline[1pt]
  Copulas & Copula-ResLogit  & & \ \ \ \ & &  Copula-Logit  \\
\cline{2-9}
        &   $LL(\omega)$ & Parameters & AIC & $\theta$ &\ \ \ \ $LL(\lambda)$ & Parameters & AIC & $\theta$\\
\hline[1pt]
    \textbf{Frank}  &    -339.44 & 167 & 1012.88 & -0.715 & -785.35 & 31 & 1632.71 & -0.613 \\
    \textbf{FGM}  &    -339.41 & 167 & 1012.83  & -0.387 & -785.27 & 31 & 1632.54 & -0.327 \\
    \textbf{Gaussian}  &   -341.03 & 167 & 1016.07 & -0.252 & -786.044 & 31 & 1634.08 & -0.122 \\
    \textbf{AMH}  &    -339.36 & 167 & 1012.73  & -0.471 & -785.133 & 31 & 1632.26& -0.385  \\
    \textbf{Product}  &    -340.30 & 166 & 1012.60  & 0 & -786.349 & 30 & 1632.69& 0  \\
\hline[1pt]
\end{tblr}\
\end{table}

In the AMH-Logit, the significant copula parameter indicates a negative dependency between pedestrian stress levels and their waiting times rooted in unobserved variables. The results suggest that there may be unobserved confounders that increase pedestrians' stress levels while possibly influencing them to wait for shorter periods on the sidewalk, or vice versa. These unobserved variables can be attributed to pedestrians' attitudes and beliefs. For instance, pedestrians' attitudes towards mixed traffic conditions may have opposite effects on stress levels and waiting times. Concerning stress, the recent widespread awareness and trust in artificial intelligence technologies, particularly the driving assistance services and autonomous vehicle braking systems, fostered by social media advertisements, can decrease pedestrians' stress in crossing. In contrast, the perception of mixed traffic as an unfamiliar environment can significantly decrease pedestrians' action to cross immediately. Another example can be pedestrians' behaviour in a group. Group crossing could potentially mitigate pedestrians' stress levels, yet may also increase waiting time levels as individuals adhere to collective decisions \cite{wilbrink2021impact}. It is worth mentioning again that this association between pedestrian stress levels and their waiting times is rooted in common unobserved variables and analysis of the direct association between pedestrian crossing components (stress level and wait time) requires further investigation employing causal learning and causal inference methodologies. The significant finding of this analysis is the existence of common unobserved confounders among pedestrian crossing components, leading to confounding associations between them.

In contrast to the traditional copula model, in the Copula-ResLogit, the independent structure (Product copula function) outperforms other copula functions. This result highlights the deep learning component’s ability to deal with unobserved confounding impacts. In fact, by measuring unobserved variables and conditioning on them in the model, the unobserved confounding dependencies are blocked, allowing the proposed model to eliminate associations between dependent variables rooted in common unobserved variables. It is important to note that in some case studies, common unobserved confounders may still persist within the random error term, potentially leading to correlations between dependent variables. The analysis in the pedestrian crossing dataset shows that the proposed model effectively adjusts for unobserved confounders, successfully rendering the two dependent variables uncorrelated.

Table \ref{ModelCoefficientResults-pedestrian} shows the results of estimated parameters of significant observed variables for both Product-ResLogit and AMH-Logit models. In general, a positive coefficient indicates that the associated variables lead to higher levels of stress and wait time, whereas negative parameters suggest the opposite effect. Regarding pedestrians' characteristics, the findings indicate a significant relationship between gender and pedestrian crossing behaviour, revealing that women tend to exhibit higher stress levels and a greater inclination to wait longer for optimal crossing opportunities. In addition, individuals who frequently utilize public transportation, walking, or cycling are prone to experiencing higher stress. Furthermore, it is observed that users of public transportation typically endure shorter waiting times compared to those utilizing private vehicles. This phenomenon can be attributed to the structured scheduling of public transport systems, which compels commuters to adhere to predetermined departure times. This result aligns with the previous studies such as \citep{kamal2024ordinal, kalatian2021decoding}. In the analysis of road attributes, the results show that pedestrians may encounter lower stress levels on one-way roads, whereas waiting times tend to be shorter on two-way roads. Furthermore, wider lane widths appear to correlate with increased pedestrian waiting times. In addition, findings highlight the influence of traffic density on pedestrian wait times, with higher traffic density corresponding to longer waiting times. This suggests that higher traffic density, indicative of more challenging conditions for pedestrians to identify suitable crossing opportunities, contributes to longer waiting times. Regarding traffic conditions, the results demonstrate that with the introduction of AVs into urban roads, pedestrians probably will feel less stress levels compared to the current traffic conditions and wait longer on the sidewalk. In terms of stress level, this outcome is particularly understandable given the general trust in AV technology. The virtual reality dataset enables analysis of the impact of environmental conditions on pedestrian crossing behaviour. The results indicate that snowy weather increases both stress levels and waiting times. This can be attributed to the reduced visibility experienced by pedestrians in such weather conditions, limiting their ability to assess crossing opportunities effectively. Moreover, findings suggest a decreased likelihood of longer waiting times on sidewalks during nighttime hours. This phenomenon could be attributed to factors such as reduced vehicular traffic or enhanced visibility due to artificial lighting. These results are consistent with \citep{kamal2024ordinal}.

\begin{table}[!htbp]
\scriptsize
\centering
\caption{Results of pedestrian stress and waiting time Copula joint models}
\begin{tblr}{
  colspec = {lcccc},
  cell{1}{2,4} = {c=2}{c}, 
}
\hline[1pt]
  Variable & Product-ResLogit  &  \ \  &    AMH-Logit  \\
\cline{2-5}
        &    Stress Level & Wait  Time & \ \ \ Stress Level & Wait  Time \\
                &    Est. (Std. Err.) & Est. (Std. Err.) & \ \ \ Est. (Std. Err.) & Est. (Std. Err.) \\
\hline[1pt]
\emph{\textbf{Socio-demographic}} \\
	Age 30-39  &  0.191(0.008) & -1.754(0.030) & 0.495(0.246) & -1.421(0.206) \\
 	Age 40-49  &  - & - & 2.678(0.857) & -1.052(0.718) \\
	Age over 50  &   1.949(0.018) & - &  2.187(0.353) & -0.492(0.340) \\
	Female  &  0.539(0.012) & 0.985(0.031) & 0.834(0.255) & 0.972(0.216) \\
	Driving license  &   0.378(0.014) & -0.957(0.047) & 0.498(0.461) & - \\
	One car  &  0.352(0.011) & 1.189(0.030) & 0.335(0.285) & 0.790(0.208) \\
        Over one car  &  0.095(0.008) & 0.234(0.029) & 0.351(0.279) & - \\
	Active mode  &  0.283(0.010) & -0.235(0.031) & 0.545(0.322) & -0.315(0.248) \\
	Public mode  &  0.477(0.009) & -0.973(0.031) & -0.893(0.257) & -0.855(0.228) \\
	\hline
    \emph{\textbf{Road attributes}} \\
    Low lane width  &  -0.113(0.007) &  -0.271(0.025) & - & -0.226(0.213)\\
    High lane width  &  -0.230(0.008) & 0.728(0.027) & -0.642(0.220) & 0.362(0.225) \\
    Two-way with a median  &  0.243(0.008) & -1.093(0.025) & 0.586(0.232) & -0.519(0.186) \\
    Two-way   &  0.181(0.008) & -0.474(0.026) & 0.602(0.235) & - \\
    Density  &  - & 0.048(0.001) & - & 0.030(0.012) \\
    \hline
    \emph{\textbf{Traffic condition}} \\
    Mixed traffic condition  &  -0.489(0.024) & 0.137(0.068) & - & - \\
    Fully automated condition  &  -0.596(0.018) & 1.093(0.047) & -0.351(0.387) & 0.538(0.347) \\
    \hline
    \emph{\textbf{Environmental variables}} \\
	Snowy  &  0.215(0.006) & 0.169(0.023) & 0.368(0.198) & - \\
        Night  &  0.159(0.007) & -0.345(0.021) & 0.279(0.191) & -0.237(0.181)\\
	\hline
    \hline
    Copula parameter  &   0    & &  &    -0.385(0.252)\\
    No. observation  &      1046    &  & &      1046\\
    No. parameters  &       166   & & &    31 \\
    Log-likelihood  &      -340.303   &  & &   -785.133 \\
    AIC  &   1012.60 & & & 1632.26\\
   \textbf{Mean Prediction Erorr(MPE)} &   45.86\% & & & 75.16\%\\
\hline[1pt]
\label{ModelCoefficientResults-pedestrian}
\end{tblr}\
\end{table}

\subsection{Travel mode and distance}
\label{Results-london}
To analyze the potential confounding association between travel mode and distance, different copula functions are developed. Table \ref{ModelPerformanceResults-london} presents the performance of copulas for both Copula-ResLogit and Copula-Logit. For both model structures, the Frank copula outperforms other copula functions, as indicated by lower AIC values and copula parameters estimated within acceptable ranges. 
In this case study, three copula parameters are estimated to capture the dependency between travel mode choices-active modes, private car, and public transportation- travel distance. The interpretation of copula parameters differs from the main impacts of observed variables as the copula parameters capture the influence of unobserved factors, which are still black-box in this study. The results of the Frank-ResLogit model indicate a positive dependency between private car use and longer travel distances, suggesting the presence of common unobserved factors that simultaneously increase the likelihood of choosing a private car and choosing a farther destination. In other words, certain unobserved variables influence both decisions in the same direction, leading to a positive correlation between them. Conversely, the copula parameter for public transit is negative, indicating that unobserved factors captured in the error terms induce a negative dependence between the two dependent variables—public transit usage and travel distance. For instance, for non-mandatory trips, comfort-related considerations (e.g., avoidance of parking difficulties) may encourage individuals to choose public transit. However, such preferences for comfort may also be associated with a preference for shorter trips. In addition, for active modes, the reasonable results reveal a negative confounding association between the choice of active travel and travel distance.

Moreover, the results indicate that even after accounting for unobserved confounders through the Copula-ResLogit model, an unobserved dependence remains between travel mode and travel distance. This implies that the ResNet block does not fully capture all unobserved factors, and that some unobserved confounders continue to create confounding associations between the two variables. This issue may be addressed by increasing residual depth in ResNet-based model structures. Because, unlike vanilla deep neural networks, where increasing depth can lead to performance degradation, previous studies have shown that the performance of ResNet-based architectures improves with depth of residual layers \cite{kamal2024ordinal, wong2021reslogit}. To further investigate, our study examines whether increasing the number of residual layers affects the model’s ability to mitigate the influence of unobserved confounders. Table \ref{residuallayersimpact} compares the performance of models with 16 and 32 residual layers. It is important to note that the only structural difference between these models is the number of residual layers. A key finding is that by increasing the number of residual layers, the Product-ResLogit model outperforms the Frank-ResLogit model. This result suggests that additional residual layers enable the model to capture remaining unobserved confounding. With 32 residual layers, the model significantly removes the confounding association between travel mode and travel distance, rendering them conditionally independent.

\begin{table}[!htbp]
\footnotesize
\centering
\caption{Model Performance of Copula-ResLogit and Copula-Logit for Travel Mode and Distance Analysis}
\label{ModelPerformanceResults-london}

\begin{tabular}{lcccccccc}

\hline
\tiny
\textbf{Copulas} 
& \multicolumn{4}{c}{\tiny\textbf{Copula-ResLogit}} 
& \multicolumn{4}{c}{\tiny\textbf{Copula-Logit}} \\
\cline{2-5} \cline{6-9}
& \textbf{\tiny LL} 
& \textbf{\tiny\#Params} 
& \textbf{\tiny AIC} 
& $\boldsymbol{\tiny\theta}$ 
& \textbf{\tiny LL} 
& \textbf{\tiny\#Params} 
& \textbf{\tiny AIC} 
& $\boldsymbol{\tiny\theta}$ \\
\hline
\textbf{\tiny Frank}     & -28977.95  & 239 & 58433.90 & (-0.66, 0.13, -5.70) & -45614.78   & 28 & 91285.56 &  (0.64,  -8.12, -12.01) \\
\textbf{\tiny FGM}       & -40652.66  & 239 & 81783.32 & $\theta \notin [-1, 1]$  &-46796.30   & 28 & 93648.60 & $\theta \notin [-1, 1]$\\
\textbf{\tiny Gaussian}  & -29650.21  & 239 & 59778.42 &  $\theta \notin [-1, 1]$ &-45743.55  & 28 & 91543.10 & $\theta \notin [-1, 1]$ \\
\textbf{\tiny AMH}       & -29645.31  & 239 & 59768.62 & $\theta \notin [-1, 1]$ &-45650.55   & 28 & 91357.10 & $\theta \notin [-1, 1]$ \\
\textbf{\tiny Product}   & -29553.56  & 236 & 59579.12 & (0, 0, 0)  & -48527.24   & 31 & 97104.48 & (0, 0, 0) \\
\hline
\end{tabular}
\end{table}

\begin{table}[!htbp]
\footnotesize
\centering
\caption{Model Performance of Copula-ResLogit with Different Residual Layers}
\label{residuallayersimpact}

\begin{tabular}{lcccc}
\hline
\textbf{Copula} 
& \multicolumn{2}{c}{\textbf{ResLogit (16 layers)}} 
& \multicolumn{2}{c}{\textbf{ResLogit (32 layers)}} \\
\cline{2-3} \cline{4-5}
& \textbf{LL} & \textbf{AIC} 
& \textbf{LL} & \textbf{AIC} \\
\hline
\textbf{Frank}   & -28977.95 & 58433.90 & -29783.83 & 60333.66 \\
\textbf{Product} & -29553.56 & 59579.12 & -29651.76 & 60057.52 \\
\hline
\end{tabular}
\end{table}

The comparative analysis highlights the critical role of tuning the depth of residual layers in addressing unobserved confounding in causality-based studies. In other words, the copula functions measure the level of dependency for both model settings and indicate that by adding more residual layers, the model’s capacity increases for eliminating unobserved confounders. However, increasing model depth inherently leads to greater model complexity. Therefore, in studies where the primary focus is on model flexibility and predictive performance, rather than what-if causal analysis, the optimal number of residual layers should be selected based on model performance metrics.

Table \ref{ModelCoefficientResults-london} shows the results of significant parameters estimated for observed variables for both Product-ResLogit and Frank-Logit models. Regarding the influence of the estimated parameters, our findings indicate that increased household access to private vehicles is associated with a higher likelihood of choosing private cars, while simultaneously showing a negative effect on the probability of choosing public transportation. Moreover, higher levels of car ownership are linked to a decreased likelihood of selecting more distant destinations. These findings are consistent with previous research using this dataset \cite{kamal2024ordinal}. A similar pattern is observed with respect to holding a valid driver’s license, which is positively associated with car usage and negatively associated with the choice of longer travel distances. In general, the results demonstrate that as individuals age, their preference for both private car and public transportation decreases compared to active modes, including walking and cycling. Regarding travel distance, older individuals are also more likely to choose destinations closer to their home location. In addition, women are more likely to choose active modes for non-mandatory trips compared to private car and tend to select shorter travel distances. Another finding shows the negative impact of travel cost on mode choices. In fact, as travel costs increase, individuals' likelihood of choosing a given mode decreases. In addition, increasing transit cost negatively impacts the travel distance of a trip, meaning that people prefer to choose a closer destination.

\begin{table}[!htbp]
\scriptsize
\centering
\caption{Results of Travel Mode and Distance Copula Joint Models}
\label{ModelCoefficientResults-london}

\begin{tabular}{lcccccc}
\hline
\multirow{2}{*}{\textbf{Variable}} & \multicolumn{3}{c}{\textbf{Product-ResLogit}} & \multicolumn{3}{c}{\textbf{Frank-Logit}} \\
                                   & Private Car & Public Transit & Travel Distance & Private Car & Public Transit & Travel Distance \\
\hline
\multicolumn{7}{l}{\emph{\textbf{Socio-demographic}}} \\
Age 30–45              & -0.034 (0.002) & -0.294 (0.017) & -- & -0.206 (0.011) & -0.210 (0.011) & -0.263 (0.014) \\
Age 45–60              & -0.047 (0.002) & -0.852 (0.021) & -0.030 (0.003) & -0.053 (0.011) & -0.286 (0.013) & -0.271 (0.014) \\
Age over 60            & -0.112 (0.002) & -0.136 (0.005) & -0.041 (0.002) & 0.015 (0.011) & 0.862 (0.011) & 0.622 (0.012) \\
Female                 & -0.045 (0.001) & 0.038 (0.005) & -0.018 (0.002) & 0.173 (0.007)  & 0.196 (0.008) & -0.138 (0.009) \\
Disability             &  0.091 (0.003) & -0.957 (0.047) &       --       &      -0.739 (0.020)       & -- & -- \\
Driving License        &  0.034 (0.001) & -2.564 (0.012) & -0.040 (0.002) &  1.094 (0.009)       & -0.500 (0.010) & -0.058 (0.011) \\
Unrestricted Car Access&  0.043 (0.002) & -1.690 (0.019) & -0.071 (0.002) & 0.741 (0.010)  & -0.395 (0.013) & 0.400 (0.011) \\
\hline
\multicolumn{7}{l}{\emph{\textbf{Travel Attributes}}} \\
Transit Cost           &       --       & -0.235 (0.005) &  0.007 (0.001) &      --        & 0.270 (0.003) & 0.808 (0.006) \\
Driving Cost           & -1.467 (0.004) &       --       &  2.028 (0.005) & -0.173 (0.002)  & -1.754 (0.030) & 0.309 (0.003) \\
\hline
\hline
\multicolumn{7}{l}{\emph{\textbf{Model Fit Statistics}}} \\
Copula Parameter       & \multicolumn{3}{c}{(0, 0, 0)} & \multicolumn{3}{c}{(0.79, -5.42, -8.46)} \\
No. of Observations    & \multicolumn{3}{c}{45,547} & \multicolumn{3}{c}{45,547}\\
No. of Parameters      & \multicolumn{3}{c}{239} & \multicolumn{3}{c}{28}\\
Log-Likelihood         & \multicolumn{3}{c}{-29651.76} & \multicolumn{3}{c}{-45614.78}\\
AIC                    & \multicolumn{3}{c}{60057.52} & \multicolumn{3}{c}{91285.56}\\
Mean Prediction Error (MPE) & \multicolumn{3}{c}{39.54\%} & \multicolumn{3}{c}{55.27\%}\\
\hline
\end{tabular}
\end{table}

\section{Conclusion}
This study focuses on the potential influence of unobserved confounders in the decision-making process, particularly in what-if scenario analyses where the emphasis is on identifying direct and causal relationships between variables. It aims to emphasize the potential presence of unobserved variables that create non-causal dependencies between factors. To capture and control this unobserved dependency, the study proposes a novel deep learning-based joint model structure using the copula approach, called the Copula-ResLogit model. The proposed framework leverages the ability of deep learning components in Residual Neural Network (ResNet)-based models to address the impact of unobserved confounders. In addition, the formulation of this model utilizes a copula joint structure to measure the dependency between variables. In other words, this research study enables modellers to first investigate unobserved confounding associations using traditional joint modelling approaches and then address these associations by incorporating deep learning components. It also assesses whether any unobserved associations persist after adjusting for unobserved confounders. In other words, the proposed model is still able to capture the potential unobserved confounders, that may not be measured by the deep learning components. Copula-ResLogit can be classified as a logical extension of ResNet-based models, explicitly designed for dealing with non-causal dependency between factors.

In this study, two primary components of the pedestrian crossing decision process within the context of Automated Vehicles (AVs) are investigated jointly. To analyze this futuristic scenario in the presence of AVs, a Virtual Reality (VR) dataset is utilized. This study explicitly analyzes the non-causal association between two ordinal crossing components: pedestrian stress levels and waiting times, stemming from common unobserved variables. In addition, a multinomial-ordinal Copula-ResLogit model is developed to examine any non-causal dependency between travel mode choice and travel distance using London travel behaviour data. 

The study compares the hybrid bivariate joint model structure with a completely theory-driven copula-based joint model called Copula-Logit. for both case studies, four prevalent copula functions which are capable of flexibly capturing both negative and positive dependencies, are used in Copula-ResLogit and Copula-Logit. Furthermore, the performance of the developed joint frameworks is examined against the corresponding independent model, where no confoundingg correlation between dependent variables is assumed. In other words, this analysis allows for the examination of whether a non-causal confounding path exists between the two dependent variables before and after accounting for unobserved confounders.

This research study utilizes various copulas to examine unobserved associations between dependent variables. In both datasets, the Copula-Logit models reveal non-causal dependency between dependent variables. This finding shows that there are unobserved confounders that create the non-causal association between the factors. For instance, the analysis of copula parameters in the Copula-Logit model shows a negative relationship between pedestrian stress levels and waiting times. However, the analysis of the developed Copula-ResLogit model structure indicates that the independent model structure outperforms other copula functions. This finding suggests that the inclusion of residual layers in the Copula-ResLogit model enables modellers to effectively eliminate the unobserved dependencies stemming from unobserved confounders. Interestingly, the case study on travel mode choice and travel distance suggests that some unobserved confounders may remain unaccounted for by the ResNet block. However, increasing the number of residual layers appears to allow the model to fully mitigate these confounding effects. In the context of causal analysis, addressing any confounding effects between variables is crucial, as the primary focus should be on direct causal relationships and our results on both case studies highlight that our proposed model is able to completely isolate causal association from non-causal unobserved association.

It is important to acknowledge that while the Copula-ResLogit model has made significant strides in causal analysis, the proposed model primarily concentrates on common unobserved confounders. However, causality-based analysis focuses on causal dependencies between observed variables after adjusting for both observed and unobserved confounders. Therefore, additional research is needed to explore other types of confounders that could create confounding paths between these variables. Building on our study, a future direction can be a novel combination of causal inference techniques and copula-joint modelling for isolating causal associations from both observed and unobserved associations between variables.

\section{Acknowledgements}
This research was funded by a grant from the Canada Research Chair program in Disruptive Transportation Technologies and Services (CRC-2017-00038) and NSERC Discovery (RGPIN-2020-04492) fund.

\bibliographystyle{trb}
\bibliography{trb_template}

\end{document}